\documentclass{article}
\pdfoutput=1 

\usepackage{arxiv}

\usepackage{graphicx}
\usepackage{url}
\usepackage{threeparttable}
\usepackage{multirow}
\usepackage{amsmath}
\usepackage{amssymb}
\usepackage{bm}
\usepackage{enumitem}
\usepackage{algorithm}  
\usepackage{algorithmic}
\usepackage{subfigure}
\usepackage{booktabs}
\usepackage{color}
\usepackage{microtype}

\usepackage{colortbl}

\usepackage{soul, color, xcolor}
\soulregister{\cite}7 
\soulregister{\ref}7 
\soulregister{\url}7 
\soulregister{\footnote}7 

\usepackage{array}
\makeatletter
\newcommand{\thickhline}{%
    \noalign {\ifnum 0=`}\fi \hrule height 1pt
    \futurelet \reserved@a \@xhline
}
\newcolumntype{"}{@{\hskip\tabcolsep\vrule width 1pt\hskip\tabcolsep}}
\makeatother

\makeatletter
\renewcommand{\maketag@@@}[1]{\hbox{\m@th\normalsize\normalfont#1}}%
\makeatother

\usepackage{diagbox}
\usepackage{amsfonts}
\usepackage{makecell}

\usepackage[utf8]{inputenc} 
\usepackage[T1]{fontenc}    
\usepackage{hyperref}       
\usepackage{url}            
\usepackage{booktabs}       
\usepackage{amsfonts}       
\usepackage{nicefrac}       
\usepackage{microtype}      
\usepackage{cleveref}       
\usepackage{lipsum}         
\usepackage{graphicx}
\usepackage{natbib}
\usepackage{doi}

\title{Large Language Models for Constructing and Optimizing Machine Learning Workflows: A Survey\thanks{This paper is under review. © © 20xx IEEE. Personal use of this material is permitted. Permission from IEEE must be obtained for all other uses, in any current or future media, including reprinting/republishing this material for advertising or promotional purposes, creating new collective works, for resale or redistribution to servers or lists, or reuse of any copyrighted component of this work in other works.}}

\date{}

\newif\ifuniqueAffiliation

\ifuniqueAffiliation 

\else
\usepackage{authblk}

\author[1]{%
	{Yang Gu\thanks{gu\_yang@sjtu.edu.cn. Corresponding author is Jian Cao (cao-jian@sjtu.edu.cn). For recent work on LLMs for ML workflows, visit our repository: \url{https://github.com/t-harden/LLM4AutoML}.}}%
}
\author[1]{%
	{Hengyu You}%
}
\author[1]{%
	{Jian Cao}%
}
\author[2]{%
	{Muran Yu}%
}
\author[1]{%
	{Haoran Fan}%
}
\author[1]{%
	{Shiyou Qian}%
}
\affil[1]{Shanghai Jiao Tong University, Shanghai, China}
\affil[2]{Stanford University, Stanford, California, USA}
\fi

\renewcommand{\shorttitle}{LLMs for Constructing and Optimizing ML Workflows}

\hypersetup{
pdftitle={A template for the arxiv style},
pdfsubject={q-bio.NC, q-bio.QM},
pdfauthor={David S.~Hippocampus, Elias D.~Striatum},
pdfkeywords={First keyword, Second keyword, More},
}

\begin{document}
\maketitle

\begin{abstract}
Building effective machine learning (ML) workflows to address complex tasks is a primary focus of the Automatic ML (AutoML) community and a critical step toward achieving artificial general intelligence (AGI). Recently, the integration of Large Language Models (LLMs) into ML workflows has shown great potential for automating and enhancing various stages of the ML pipeline. This survey provides a comprehensive and up-to-date review of recent advancements in using LLMs to construct and optimize ML workflows, focusing on key components encompassing data and feature engineering, model selection and hyperparameter optimization, and workflow evaluation. We discuss both the advantages and limitations of LLM-driven approaches, emphasizing their capacity to streamline and enhance ML workflow modeling process through language understanding, reasoning, interaction, and generation. Finally, we highlight open challenges and propose future research directions to advance the effective application of LLMs in ML workflows.
\end{abstract}


\section{Introduction}
\label{1}
In the era of big data, machine learning (ML) workflows have become essential across various sectors for processing and analyzing large-scale data \cite{xin2021whither, nikitin2022automated}. To support the development and sharing of ML workflows, numerous repositories have been established, showcasing diverse paradigms for data analysis. For instance, KNIME offers a repository with over 25,000 workflows and 2,200 components \cite{ordenes2021machine}, providing a comprehensive collection of rigorously tested, practical models complete with detailed specifications. However, despite the availability of these resources, manually constructing and optimizing workflows to meet complex task requirements remains a knowledge-intensive and time-consuming challenge for most people.

The advent of Large Language Models (LLMs) has recently revolutionized artificial intelligence (AI) and ML, delivering advanced capabilities in natural language understanding and generation \cite{hollmann2024large, wang2024grammar}. Models such as OpenAI's GPT-4 \cite{achiam2023gpt} and Meta AI's LLaMA-3 \cite{touvron2023llama} have demonstrated exceptional performance across a wide range of natural language processing (NLP) tasks, thanks to their extensive training on large-scale text datasets. Additionally, multimodal LLMs \cite{hu2024bliva, tai2024link, luo2024autom3l}, which incorporate various data types like audio and images, allow for richer interactions by processing and generating non-textual information. Their impressive capabilities have led to widespread adoption across multiple domains \cite{gu2023plan, klievtsova2023conversational, zhang2023multimodal}. As ML tasks and workflows become increasingly complex, often involving diverse modalities and domains, the potential of LLMs to automate and enhance these workflows has garnered significant attention from the research community \cite{xiao2024verbalized, hong2024data}. This progress is also viewed as a pivotal step toward achieving artificial general intelligence (AGI).

\begin{figure}[htbp]
  \centering
  \includegraphics[width=1\textwidth]{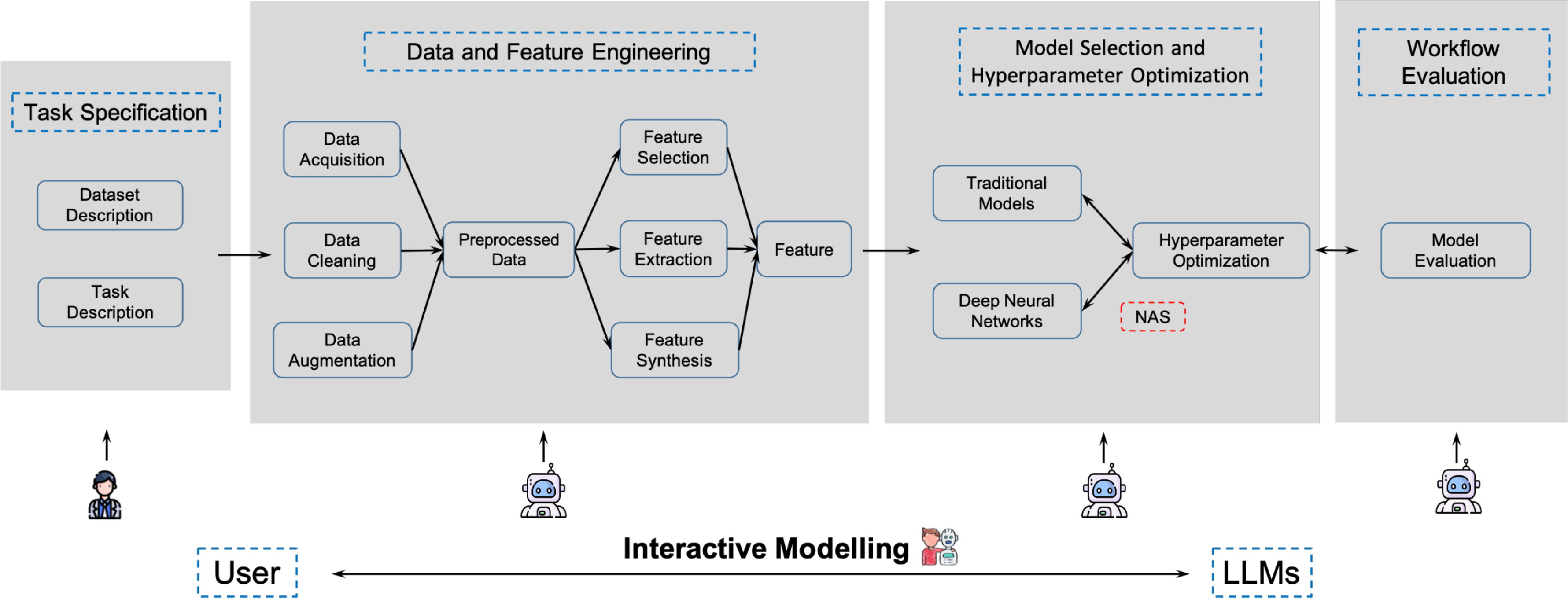}
  \caption{An overview of the Machine Learning Workflow, where task specification serves as the input, encompassing key stages of data and feature engineering, model selection and hyperparameter optimization, and workflow evaluation.}
  \label{fig-MLWorkflow}
\end{figure}

As illustrated in Fig. \ref{fig-MLWorkflow}, an ML workflow, receiving the input of task specification, typically involves a sequence of interconnected steps \cite{de2022automating}, including data and feature engineering, model selection and hyperparameter optimization, and workflow evaluation. Traditionally, these steps demand considerable manual effort and specialized domain expertise, which can limit scalability and adaptability, particularly when working with large, high-dimensional datasets \cite{lazebnik2022substrat}. This challenge has driven the development of automated machine learning (AutoML), which seeks to streamline the ML workflow by automating essential modeling and optimization processes \cite{hutter2019automated}.

Despite the significant advancements brought by AutoML \cite{nikitin2022automated, an2023diffusionnag}, traditional AutoML frameworks still face notable challenges. First, the iterative nature of the search processes in AutoML—often involving exhaustive model selection and hyperparameter tuning—can be extremely time-consuming and computationally expensive \cite{olson2016tpot}. Second, these methods typically struggle to leverage valuable historical and human knowledge from diverse sources effectively. Even with the integration of meta-learning and Bayesian optimization techniques \cite{feurer2022auto, saha2022sapientml}, traditional AutoML systems often lack the ability to collaborate seamlessly with human experts, thereby limiting their flexibility and adaptability. Finally, the models generated by AutoML are often criticized for their lack of interpretability, making it difficult for practitioners to understand and trust the decisions made by these systems \cite{shah2021towards, zhang2023mlcopilot}.

These limitations highlight the need for more advanced solutions, where LLMs demonstrate significant potential by incorporating prior knowledge, facilitating human-AI collaboration, and producing more interpretable outcomes. For instance, in hyperparameter optimization \cite{zhang2023using, liu2024large}, LLMs can leverage historical data and domain-specific insights to predict optimal configurations, thereby enhancing model performance and reducing the reliance on exhaustive trial-and-error methods. Furthermore, the sophisticated NLP capabilities of LLMs enable them to act as interactive agents or chatbots, capable of generating and adjusting AutoML code based on contextual information provided by users \cite{ arteaga2024support, guo2024ds, dakhel2023github}. Crucially, in the process of conversational construction of ML workflows—such as feature engineering \cite{hollmann2024large}—LLMs have shown the ability to generate human-readable and explainable features, offering a level of transparency and guidance that surpasses many traditional black-box AutoML systems \cite{zhang2024dynamic, nam2024optimized}.

However, integrating LLMs into ML workflows also presents several challenges. Issues such as reasoning hallucinations, ethical concerns, and the substantial computational demands of deploying large-scale models remain significant barriers to their widespread adoption \cite{bommasani2021opportunities, hollmann2024large, yao2024survey}. Additionally, while LLMs have demonstrated impressive results in specific tasks, their effectiveness across the entire spectrum of ML workflow stages \cite{zhang2023mlcopilot, wecoai_wecoaiaideml_2024} still requires thorough investigation and evaluation using some benchmarks like MLE-Bench \cite{chan2024mle} and MLAgentBench \cite{huang2024mlagentbench}.

This survey provides a comprehensive overview of the current state of research on the application of LLMs in constructing and optimizing ML workflows. To the best of our knowledge, it is the first survey to systematically address every stage of the ML workflow (Fig. \ref{fig-MLWorkflow}), distinguishing it from previous reviews, such as \cite{tornede2023automl}, which primarily explore the broader opportunities of an integration of LLMs and AutoML. We organize recent papers based on the specific steps in which LLMs have been utilized, as summarized in Table \ref{tab-summary}, with entries sorted by publication year (oldest to newest) and the order of workflow steps. Notably, our focus is on capturing the breadth and possibility of research within the specific context of constructing and optimizing ML workflows, rather than providing an exhaustive examination of all LLM-related methods at each individual stage.

\begin{table}[hpt]
\renewcommand\arraystretch{1.15}
  \caption{Comparison of existing methods that utilize LLMs for constructing and optimizing ML workflows in terms of specific workflow components.}
  \label{tab-summary}
  \centering
  \setlength{\tabcolsep}{2.0mm}{
  \footnotesize
    \begin{tabular}{c|c|c|c|c|c}
    \thickhline
    \textbf{Method} & \textbf{\makecell[c]{Data \\ Preprocessing}} & \textbf{\makecell[c]{Feature \\ Engineering}} & \textbf{\makecell[c]{Model \\ Selection}} & \textbf{\makecell[c]{Hyperparameter \\ Optimization}} & \textbf{\makecell[c]{Workflow \\ Evaluation}} \\ \hline
    
    LMPriors \cite{choi2022lmpriors} & & \checkmark & & & \\ \hline
    ZAP \cite{ozturk2022zero} & & & \checkmark & & \\ \hline
    
    VIDS \cite{hassan2023chatgpt} & \checkmark & & & & \\ \hline
    JarviX \cite{liu2023jarvix} & \checkmark & & & & \\ \hline
    Aliro \cite{choi2023aliro} & \checkmark & & & & \\ \hline
    AutoML-GPT \cite{zhang2023automl} & \checkmark & & \checkmark & \checkmark & \checkmark \\ \hline
    MLCopilot \cite{zhang2023mlcopilot} & & & \checkmark  & & \\ \hline
    GENIUS \cite{zheng2023can} & & & \checkmark & \checkmark & \\ \hline
    GPT-NAS \cite{yu2023gpt} & & & \checkmark & \checkmark & \\ \hline
    
    AutoMMLab \cite{yang2024autommlab} & \checkmark & & \checkmark & \checkmark & \\ \hline
    AutoML-Agent \cite{trirat2024automl} & \checkmark & & \checkmark & \checkmark & \\ \hline
    AutoM$^3$L \cite{luo2024autom3l} & \checkmark & \checkmark & \checkmark & \checkmark & \\ \hline
    Text-to-ML \cite{xu2024large} & \checkmark & \checkmark & \checkmark & & \checkmark \\ \hline
    LLM-Select \cite{jeong2024llm} & & \checkmark & & & \\ \hline
    GL-Agent \cite{wei2024versatilegraphlearningapproach} & & \checkmark & \checkmark & & \\ \hline
    CAAFE \cite{hollmann2024large} & & \checkmark & & & \\ \hline
    HuggingGPT \cite{shen2024hugginggpt} & & & \checkmark & & \\ \hline
    ModelGPT \cite{tang2024modelgpt} & & & \checkmark & & \\ \hline
    VML \cite{xiao2024verbalized} & & & \checkmark & & \checkmark \\ \hline
    GE \cite{morris2024llm} & & & & \checkmark & \\ \hline
    AgentHPO \cite{liu2024large} & & & & \checkmark & \checkmark \\ \hline
    LLAMBO \cite{liu2024largelanguagemodelsenhance} & & & & \checkmark  & \\

    \thickhline
    \end{tabular} 
    \begin{tablenotes} 
	\item \small{$^\ast$ The symbol $\checkmark$ denotes that LLMs are applied to support a particular component within the method.}
    \end{tablenotes} 
   }
\end{table}

The remainder of this paper is structured as follows. We begin with a discussion of the foundational concepts and recent developments in ML workflow and LLMs (Section \ref{2}). This is followed by an exploration of the roles LLMs play in key stages of ML workflows, including data and feature engineering (Section \ref{3}), model selection and hyperparameter optimization (Section \ref{4}), and workflow evaluation (Section \ref{5}). In Section \ref{6}, we address several open challenges and outline future research directions in this field. Finally, we conclude our survey in Section \ref{7}.

\section{Preliminaries}
\label{2}

\subsection{Machine Learning Task and Workflow}
ML is a subfield of AI that focuses on the development of algorithms and statistical models that enable computers to perform specific tasks \cite{allen2020understanding}. These tasks range from classification and regression to clustering and reinforcement learning, each designed to identify patterns and make predictions based on data \cite{zhou2021machine, lecun2015deep}. The fundamental goal of ML is to create models that generalize well from training data to unseen data, effectively solving the problem at hand \cite{chung2018unknown}. The ML Workflow, a structured sequence of components \cite{oakes2024building}, is designed to systematically and repeatably achieve the task’s objective, leading to the development of a robust and effective model.

The input of an ML workflow is typically the task specification, encompassing both the dataset definition and task description. The dataset, consisting of input data (features) and output labels (for supervised tasks), serves as the foundation for training the model \cite{saha2022sapientml}. For example, the HousingPrice dataset \cite{harrison1978hedonic} may contain a CSV table with 14 columns, where the first 13 columns represent predictive features, and the final column indicates the target variable (housing price). The task description further clarifies the model’s objective and evaluation metrics \cite{zhang2023mlcopilot}, specifying goals such as predicting housing prices using the R$^2$ metric, classifying images with the $F1$-score metric, or identifying clusters within data using the ARI metric. Importantly, datasets and tasks vary widely in structure and type, encompassing diverse data formats such as numerical values, text, images, and time series \cite{huang2024mlagentbench}.

\subsection{Background on LLMs}
Large Language Models (LLMs) are advanced neural networks, primarily built on Transformer architectures \cite{vaswani2017attention}, that excel at processing and generating human-like text. Models such as OpenAI’s GPT-4 \cite{achiam2023gpt}, Google’s PaLM-2 \cite{anil2023palm}, and Meta AI’s LLaMA-3 \cite{touvron2023llama} have revolutionized natural language processing (NLP), achieving state-of-the-art performance across a wide array of tasks, including text generation, translation, summarization, and question answering \cite{minaee2024large}. Moreover, LLMs are increasingly being applied to diverse domains beyond NLP, thanks to their ability to model complex language patterns and generalize across various data types \cite{xiao2024verbalized, liu2024largelanguagemodelsenhance, yu2023bear}.

LLMs, which often contain tens to hundreds of billions of parameters \cite{zhao2023survey, wang2024survey}, are trained on vast textual datasets, enabling them to capture intricate language patterns and generate coherent, contextually accurate text. One of their most notable features is their ability to perform zero-shot and few-shot learning, where they generalize to new tasks with minimal task-specific data or examples \cite{sahoo2024systematic}. This flexibility significantly reduces the need for retraining, allowing LLMs to handle a wide range of tasks based on just a few instructions. Additionally, LLMs can break down complex tasks into intermediate reasoning steps, as demonstrated by techniques like Chain-of-Thought (CoT) \cite{wei2022chain}, Tree-of-Thought (ToT) \cite{yao2024tree}, and Graph-of-Thought (GoT) prompting \cite{besta2024graph}. LLMs can also be augmented with external knowledge sources and tools \cite{zhuang2024toolqa, fan2024survey}, enabling them to interact more effectively with users and their environment \cite{xi2023rise}. These models can be deployed as LLM-based agents, artificial entities capable of sensing their environment, making decisions, and taking actions autonomously \cite{zhao2024expel}. Furthermore, through mechanisms like reinforcement learning with human feedback (RLHF), LLMs can continually improve their performance by incorporating feedback from interactions \cite{wang2024comprehensive}.

While fine-tuning and alignment improves their performance and adds different dimensions to their abilities, there are still some important limitations that come up. Training and deploying LLMs is computationally expensive, requiring significant hardware resources, which restricts their accessibility for many organizations \cite{zhang2024llmcompass, bai2024beyond}. Another pressing concern is the phenomenon of  reasoning hallucinations, where LLMs, despite generating plausible-sounding text, can produce factually incorrect or unfaithful outputs, potentially leading to unreliable decisions in sensitive applications \cite{li2024dawn, leiser2024hill}. Furthermore, LLMs inherently operate as probabilistic models, often producing different outputs when presented with the same inputs \cite{wang2024guiding}. While parameters like temperature can be fine-tuned to control this variability, the stochastic nature of LLM responses presents challenges in ensuring consistency and reliability across use cases \cite{gruver2024large}. Moreover, the vast datasets used to train LLMs often include sensitive information, posing ethical risks around data privacy and security, especially in fields like healthcare or finance where confidentiality is paramount \cite{yao2024survey, yang2024autommlab}. These limitations are important considerations when evaluating the deployment of LLMs in real-world tasks.

In this survey, we explore the current research landscape on leveraging LLMs to construct and optimize ML workflows, with a focus on each stage of the workflow. Our goal is to provide researchers and practitioners with a comprehensive understanding of the strengths and limitations of LLMs in this context. By examining the key achievements and identifying the existing challenges, we aim for this survey to serve as a foundation for future research, fostering more effective and integrated applications of LLMs in automating ML workflows.

\section{LLMs for Data and Feature Engineering}
\label{3}
Upon receiving the task specification, the first stage in the ML workflow is data and feature engineering, which can be further divided into two key substeps: data preprocessing and feature engineering. Data preprocessing involves cleaning, transforming, and normalizing raw data to ensure consistency and quality for subsequent analysis. Feature engineering focuses on extracting informative and relevant features from the preprocessed data to enhance the performance of learning algorithms. The following subsections will explore how LLMs can support these two crucial processes.

\subsection{Data Preprocessing}
\begin{figure}[htbp]
  \centering
  \includegraphics[width=0.85\textwidth]{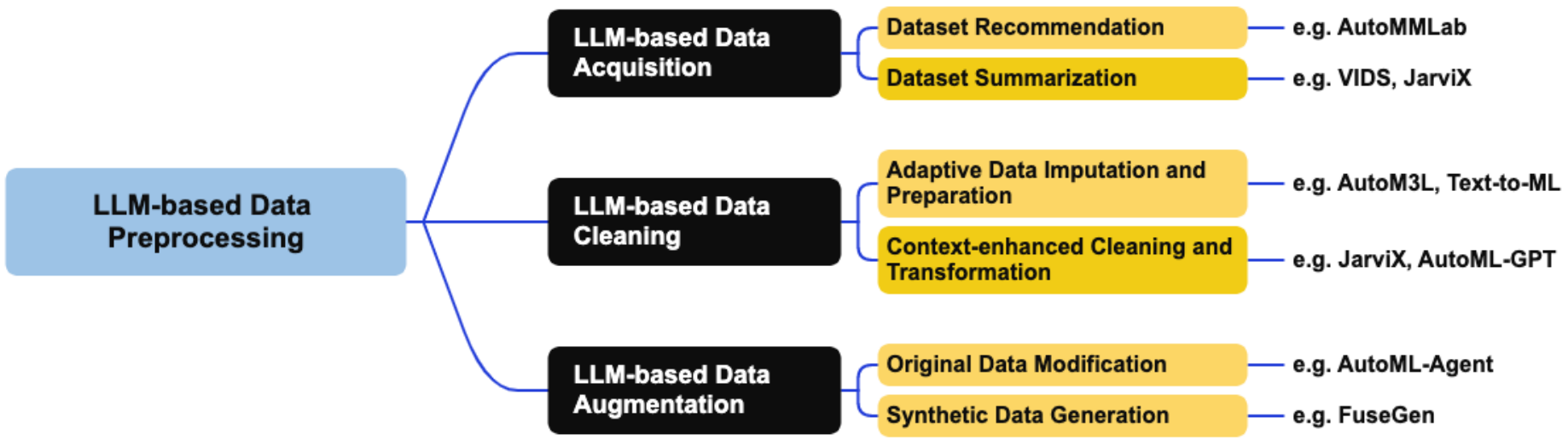}
  \caption{The main categorization of LLM-assisted Data Preprocessing methods.}
  \label{fig-DP}
\end{figure}

In many practical scenarios, the qualitative properties of raw data are not often consistent with the requirements of the target application or model \cite{zelaya2019towards, parashar2023data}. Consequently, data preprocessing has become an essential task in the machine learning application development process. Data preprocessing is typically divided into three key aspects: data acquisition, data cleaning, and data augmentation \cite{he2021automl}. Each aspect plays a crucial role in ensuring the quality and usability of the data before it is fed into machine learning models. The main categorization of LLM-assisted Data Preprocessing methods is illustrated in Fig. \ref{fig-DP}.

\begin{itemize}[leftmargin=15pt]
\item[$\bullet$] Data Acquisition - involves identifying and gathering suitable datasets.

\item[$\bullet$] Data Cleaning - filters noisy or inconsistent data to preserve the integrity of model training. 

\item[$\bullet$] Data Augmentation - enhances model robustness by artificially increasing the size and diversity of the dataset.
\end{itemize}

\subsubsection{Data Acquisition}
Data acquisition forms the foundation of the machine learning workflow by ensuring that the right data is sourced for model training and validation. This stage is critical because the quality and relevance of the collected data directly impact the final model's performance \cite{zhou2024llm}. The goal is to identify and gather datasets that align with the task at hand, which can be a time-consuming and labor-intensive process.

LLMs have emerged as powerful tools to streamline the data acquisition process by minimizing the need for extensive manual effort and domain-specific expertise. They are capable of interpreting task requirements, identifying pertinent data sources, and generating comprehensive dataset reports, thereby expediting the workflow. Approaches for LLM-assisted data acquisition generally fall into two main categories: dataset recommendation and dataset summarization.

In the context of \textbf{dataset recommendation}, some systems harness LLMs to suggest datasets that align with specific task needs. AutoMMLab \cite{yang2024autommlab} employs LLMs for natural language interaction with users, utilizing a "dataset zoo" to recommend or retrieve relevant datasets based on task descriptions. This automated selection helps tailor custom datasets to the unique requirements of specialized machine learning tasks. Similarly, AutoML-Agent \cite{trirat2024automl} employs an LLM-based Data Agent to execute sub-tasks within a decomposed plan generated through retrieval-augmented planning. These sub-tasks include dataset retrieval, pre-processing, augmentation, and analysis, ensuring that the selected datasets align seamlessly with the given task requirements. LLMs are also used for \textbf{dataset summarization}, aiding users in understanding data structure and content before further processing. The Virtual Interactive Data Scientist (VIDS) framework \cite{hassan2023chatgpt}, for example, features ChatGPT-powered agents that extract detailed insights, including dataset structures, column details, and visualization suggestions. Similarly, JarviX \cite{liu2023jarvix} incorporates an "Insight" component to automatically gather structured data information—such as column names, types, and statistics—and generate LLM-driven summary reports, giving users a quick overview of key dataset characteristics.

However, if relevant datasets are unavailable or if the LLM misinterprets task requirements, the data acquisition process can result in irrelevant or poor-quality data, potentially compromising downstream ML tasks \cite{yang2018recognition, yang2024autommlab}. Moreover, when integrating data from multiple heterogeneous sources, manual oversight may be necessary to ensure consistency and compatibility \cite{liu2023jarvix}. These limitations highlight the need for continuous refinement of LLM-driven data acquisition methods to enhance reliability and accuracy.

\subsubsection{Data Cleaning}
Data cleaning is a crucial step in the ML workflow, essential for eliminating noise, inconsistencies, and missing values that can impact model performance \cite{jesmeen2018survey, guha2024automated}. Traditionally, AutoML systems have relied heavily on rule-based approaches to handle data cleaning tasks such as error detection, imputation, and normalization \cite{neutatz2022data}. However, the introduction of LLMs into the process offers more dynamic and intelligent solutions by automating these tasks with enhanced flexibility. LLM-assisted data cleaning can be grouped into two primary categories: adaptive data imputation and preparation, and context-enhanced cleaning and transformation.

The \textbf{adaptive data imputation and preparation} category leverages LLMs to manage missing data and generate preprocessing modules that are tailored to specific tasks. For example, in the AutoM$^3$L framework \cite{luo2024autom3l}, the AFE-LLM$_{imputed}$ component uses prompts to fill in missing values based on contextual understanding, improving data completeness and ensuring datasets are ready for subsequent processing. Similarly, the Text-to-ML method \cite{xu2024large} employs LLMs to automatically generate data preparation modules, including routines for data loading and cleaning. These modules are refined iteratively through feedback, allowing the system to adapt to the unique needs of each task and continuously enhance preprocessing accuracy.

While adaptive methods focus on imputation and basic preparation, \textbf{context-enhanced cleaning and transformation} approaches aim to provide a deeper understanding of the dataset for more accurate cleaning and transformation. In this category, LLMs interpret preliminary data analyses, such as data types, correlations, and statistical summaries, to guide the cleaning process. For instance, in JarviX \cite{liu2023jarvix}, LLMs use pre-analyzed data insights stored in a database to inform cleaning recommendations, taking into account both structured and unstructured data characteristics. AutoML-GPT \cite{zhang2023automl} employs a similar strategy by using project-specific descriptions to suggest customized data transformations, recommending operations like image resizing and normalization for computer vision tasks, or tokenization and lowercasing for NLP tasks. Furthermore, the interactive data analysis tool Aliro \cite{choi2023aliro} enables users to prompt LLMs for preprocessing guidance based on observed data characteristics. For instance, if users identify outliers while exploring a PCA scatterplot, they can engage with Aliro’s chat feature to request recommendations for outlier detection methods, along with generated code snippets to effectively manage these outliers.

Despite the advantages of LLMs in identifying data patterns, they can occasionally misinterpret complex data structures, leading to incorrect or suboptimal cleaning suggestions. Moreover, handling domain-specific anomalies often requires specialized knowledge that LLMs may lack, necessitating manual intervention or targeted domain-specific training to ensure high-quality data preparation.

\subsubsection{Data Augmentation}
Data augmentation refers to techniques used to artificially increase the size of the training dataset by creating enhanced versions of existing data. This step is particularly useful for enhancing model robustness, improving generalization, and avoiding overfitting, especially in scenarios where data is scarce.

LLMs contribute significantly to data augmentation by employing techniques for modifying original data and generating synthetic datasets. In the realm of \textbf{original data modification}, LLMs assist in applying diverse transformation techniques to augment data effectively. For instance, the Data Agent in AutoML-Agent \cite{trirat2024automl} suggests various image transformation methods, such as random horizontal flipping and zooming, to enhance user-input image datasets, thereby increasing data variability and robustness for model training. For \textbf{synthetic data generation}, recent studies underscore the capabilities of pre-trained language models (PLMs) to create synthetic datasets for training target models \cite{gao2023self, yu2024large}. To address potential biases in synthetic data generated by a single PLM, FuseGen \cite{zou2024fusegen} innovatively combines multiple PLMs to collaboratively produce higher-quality synthetic datasets without incurring additional queries to individual models, thereby improving the reliability of synthetic data. Similarly, NVIDIA’s Nemotron-4 340B model family \cite{adler2024nemotron} has been integrated into synthetic data generation pipelines to produce datasets for diverse applications, including healthcare, retail, and manufacturing. Although the integration of LLMs into data augmentation is still in its early stages, their ability to modify and generate meaningful data based on learned patterns from existing datasets presents a promising avenue. This is particularly valuable in scenarios where data availability is limited, enabling more effective solutions for specific ML tasks \cite{geng2023deep}.

One challenge in using LLMs for data augmentation is the potential introduction of bias or irrelevant features into the augmented data. Additionally, ensuring that synthetic data accurately reflects real-world tasks without raising privacy or ethical concerns also remains a complex issue \cite{lin2023differentially, xie2024differentially}. 

\subsection{Feature Engineering}
While machine learning models built on deep neural architectures can automatically learn useful features \cite{bengio2013representation}, certain application settings still require explicit feature processing before model training to ensure optimal performance. Feature engineering is the process of extracting and refining relevant features from raw input data, which can significantly enhance predictive accuracy \cite{hollmann2024large}. The process typically involves three key subtopics: feature selection, feature extraction, and feature synthesis \cite{mumuni2024automated}.

\begin{itemize}[leftmargin=15pt]
\item[$\bullet$] Feature Selection - is the process of choosing the most important features that reduce feature redundancy and improve model performance by focusing on the most relevant data attributes.

\item[$\bullet$] Feature Extraction - aims to create more robust, representative and compact features by applying specific mapping functions to the raw data.

\item[$\bullet$] Feature Synthesis - involves generating new features from existing ones, creating richer representations that can better capture the underlying patterns in the dataset.
\end{itemize}

\subsubsection{Feature Selection}
Feature selection involves building a subset of features from the original set by eliminating irrelevant or redundant ones \cite{he2021automl}. This process simplifies the model, reduces the risk of overfitting, and enhances overall performance \cite{parashar2023data}. The selected features are typically diverse and highly correlated with the target variables, ensuring they contribute meaningfully to the model's predictions.

LLMs are increasingly being employed to automate feature selection by leveraging their ability to understand the semantic context of datasets. In the LMPriors framework \cite{choi2022lmpriors}, LLMs are prompted to assess whether each candidate feature should be used for predicting the target outcome. Features are selected based on the difference in log probabilities between generating a "Y" (Yes) or "N" (No) token, crossing a predefined threshold. In contrast, LLM-Select \cite{jeong2024llm} uses the generated text output directly, rather than token probabilities, to evaluate and select features. In this approach, LLMs analyze textual descriptions of features and their relationship to the target task, helping to identify the most relevant features that align with the goals of the machine learning model. This semantic-driven approach aids in selecting features that are crucial for predictive performance, thus reducing the complexity of the model. Similarly, in AutoM$^3$L \cite{luo2024autom3l}, the AFE-LLM$_{filter}$ component effectively filters out irrelevant or redundant attributes. The LLM integrates contextual information, including attributes from diverse datasets, column names from structured tables, modality inference results from MI-LLM (Modality Inference-LLM), and user instructions or task descriptions, into the prompt to enhance the feature selection process.

However, LLMs may inherit undesirable biases from their pretraining data \cite{gallegos2024bias}, which could result in biased feature selection and performance disparities across subpopulations within the dataset. This concern may be addressed by using LLM-driven feature selection in a human-in-the-loop setup or combining it with traditional data-driven methods \cite{jeong2024llm}. Additionally, there is a risk that features selected based on semantic relevance may be statistically weak, potentially diminishing their contribution to model performance.

\subsubsection{Feature Extraction}
Feature extraction is a dimensionality reduction process that transforms the original features using mapping functions to extract informative and non-redundant features based on specific metrics. Unlike feature selection, which retains the original features, feature extraction modifies them to generate new representations. Common approaches for feature extraction include methods like principal component analysis (PCA) \cite{mackiewicz1993principal}, linear discriminant analysis (LDA) \cite{xanthopoulos2013linear}, and autoencoders \cite{hinton2006reducing}.

LLMs have demonstrated significant potential in feature extraction, particularly in handling complex, multimodal datasets and generating semantically meaningful features. In the LLM-based Versatile Graph Learning approach \cite{wei2024versatilegraphlearningapproach}, LLMs are employed to select appropriate feature engineering strategies that capture both the structural and semantic properties of graph data, aligning them with the specific learning task and evaluation metric. Similarly, in the Text-to-ML method \cite{xu2024large}, LLMs automate the extraction and transformation of raw features for numerical, text, and image data by generating task-specific code. This code is then validated using automatically generated unit tests and synthetic data produced by LLMs, ensuring consistency in the extracted features across all dimensions.

A key challenge with LLM-driven feature extraction is ensuring that the generated features accurately represent the underlying data distribution. In some cases, the complexity of the dataset or domain-specific nuances may lead to suboptimal or incorrect feature extraction, which could adversely affect model performance.

\subsubsection{Feature Synthesis}
Feature synthesis involves leveraging the statistical distribution of extracted features to generate new, complementary ones. This approach is particularly useful when the existing features are insufficient to provide an adequate representation of the input data \cite{mumuni2024automated}. Traditionally, this process has relied heavily on human expertise for tasks like standardization and feature discretization. However, manually exploring all possible feature combinations is infeasible. As a result, automatic feature construction methods, such as tree-based, genetic algorithm-based and reinforcement learning-based approaches \cite{vafaie1998evolutionary, zhang2019automatic}, have been developed and have demonstrated performance comparable to or even better than human-designed features.

\begin{figure}[htbp]
  \centering
  \includegraphics[width=0.8\textwidth]{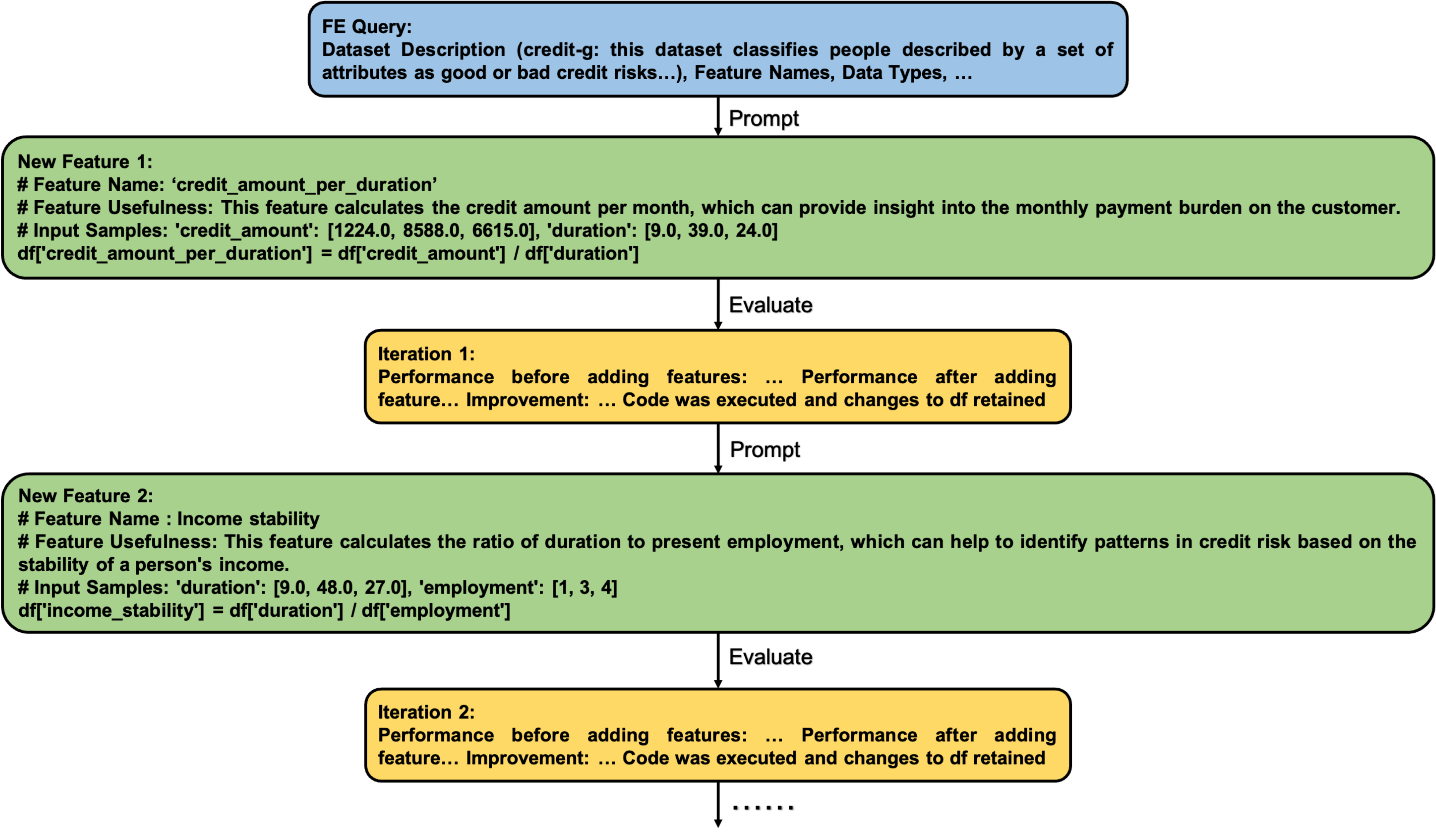}
  \caption{An example of feature synthesis by LLMs. Refer to CAAFE \cite{hollmann2024large} for detailed description.}
  \label{fig-FE}
\end{figure}

LLMs have proven highly effective in generating semantically and contextually relevant synthetic features, enhancing the feature synthesis process through an interpretable, human-in-the-loop approach. In the Context-Aware Automated Feature Engineering (CAAFE) framework \cite{hollmann2024large}, LLMs iteratively generate additional meaningful features for tabular datasets based on the dataset description, as shown in Fig. \ref{fig-FE}. This approach not only produces Python code for creating new features but also provides explanations of the utility and relevance of the generated features. CAAFE represents a step forward in semi-automated data science tasks, emphasizing the importance of context-aware solutions that extend AutoML systems into more interpretative and human-centered workflows \cite{pmlr-v235-lindauer24a}.

However, processing datasets with a large number of features can result in oversized prompts, which can be challenging for LLMs to handle efficiently. Additionally, LLMs may exhibit hallucinations, generating features and explanations that seem plausible and are logically structured but may not be grounded in the actual data. These issues highlight the need for careful oversight when using LLMs for feature synthesis.

\section{LLMs for Model Selection and Hyperparameter Optimization}
\label{4}

Model Selection (MS) and Hyperparameter Optimization (HPO) play a pivotal role in the ML workflow, directly impacting the performance and generalizability of the model. Model selection involves choosing the most suitable algorithm or model architecture for a given task \cite{yang2019oboe}, while hyperparameter optimization fine-tunes the settings that govern the behavior of the model, such as learning rate, regularization strength, and the number of layers \cite{vincent2023improved}. A related task is the Combined Algorithm Selection and Hyper-parameter optimisation (CASH). It jointly address the MS and HPO problems by treating the selection of the ML algorithm as an hyper-parameter itself \cite{thornton2013auto}.

These two processes are critical for achieving optimal model performance, as they balance the trade-offs between model complexity, training efficiency, and predictive accuracy. Effective MS and HPO ensure that models generalize well to unseen data, mitigating the risks of overfitting or underfitting. Recently, LLMs have shown promise in enhancing model selection and hyperparameter optimization by leveraging their contextual understanding and data processing capabilities \cite{zhang2023mlcopilot, xiao2024verbalized, liu2024large}. In the following subsections, we explore how LLMs are being applied to automate and improve these two crucial aspects of the ML workflow.

\subsection{Model Selection}

Model selection involves the identification of the most appropriate algorithm or model architecture for a specific task. Selected models can generally be classified into two categories: traditional models, such as decision trees \cite{quinlan1986induction} and naive Bayes classifiers \cite{rish2001empirical}, and deep neural networks (DNNs), such as convolutional neural networks (CNNs) \cite{lecun1998gradient} and recurrent neural networks (RNNs) \cite{hochreiter1997long}, which are more frequently used for tasks involving unstructured data like images and sequences. Traditionally, the model selection process relies heavily on trial and error, domain expertise, and substantial computational resources, requiring significant human effort from experts to test and compare various models \cite{lindauer2015autofolio}. The correct choice of model significantly influences predictive performance, interpretability, and scalability.

LLMs are increasingly being leveraged to streamline and automate model selection. By analyzing textual descriptions of tasks and datasets, LLMs can retrieve or generate suitable algorithms based on their pre-trained knowledge and accumulated experience, reducing the need for manual experimentation. Current methods for LLM-assisted model selection can be broadly categorized into two approaches: retrieval-based and generation-based model selection.

\begin{figure}[htbp]
    \centering

    \subfigure[Retrieval-based]{
    \includegraphics[width=3.5cm]{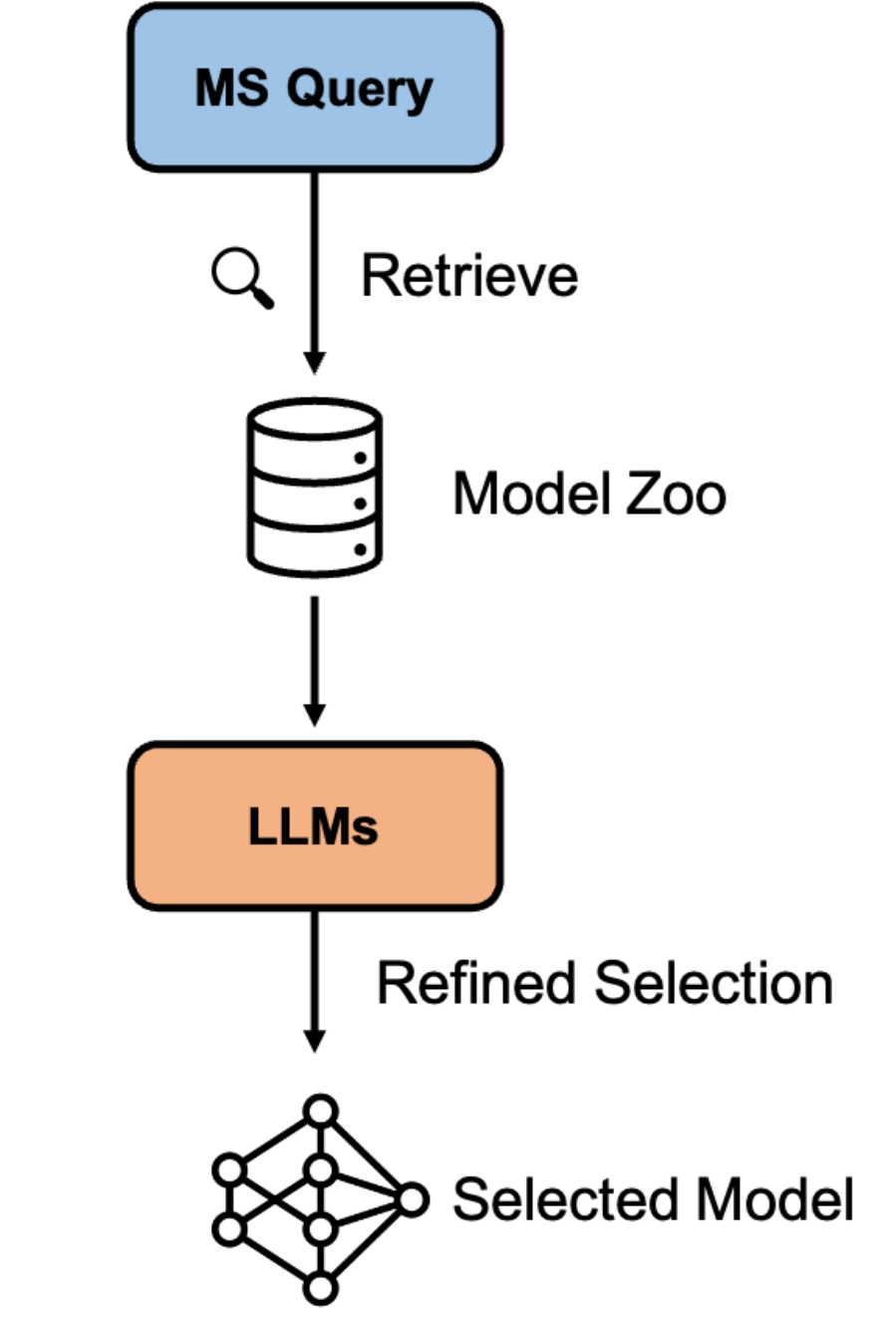}
    \label{fig-MS_re}
    }
    \quad
    \subfigure[Generation-based]{
    \includegraphics[width=3.5cm]{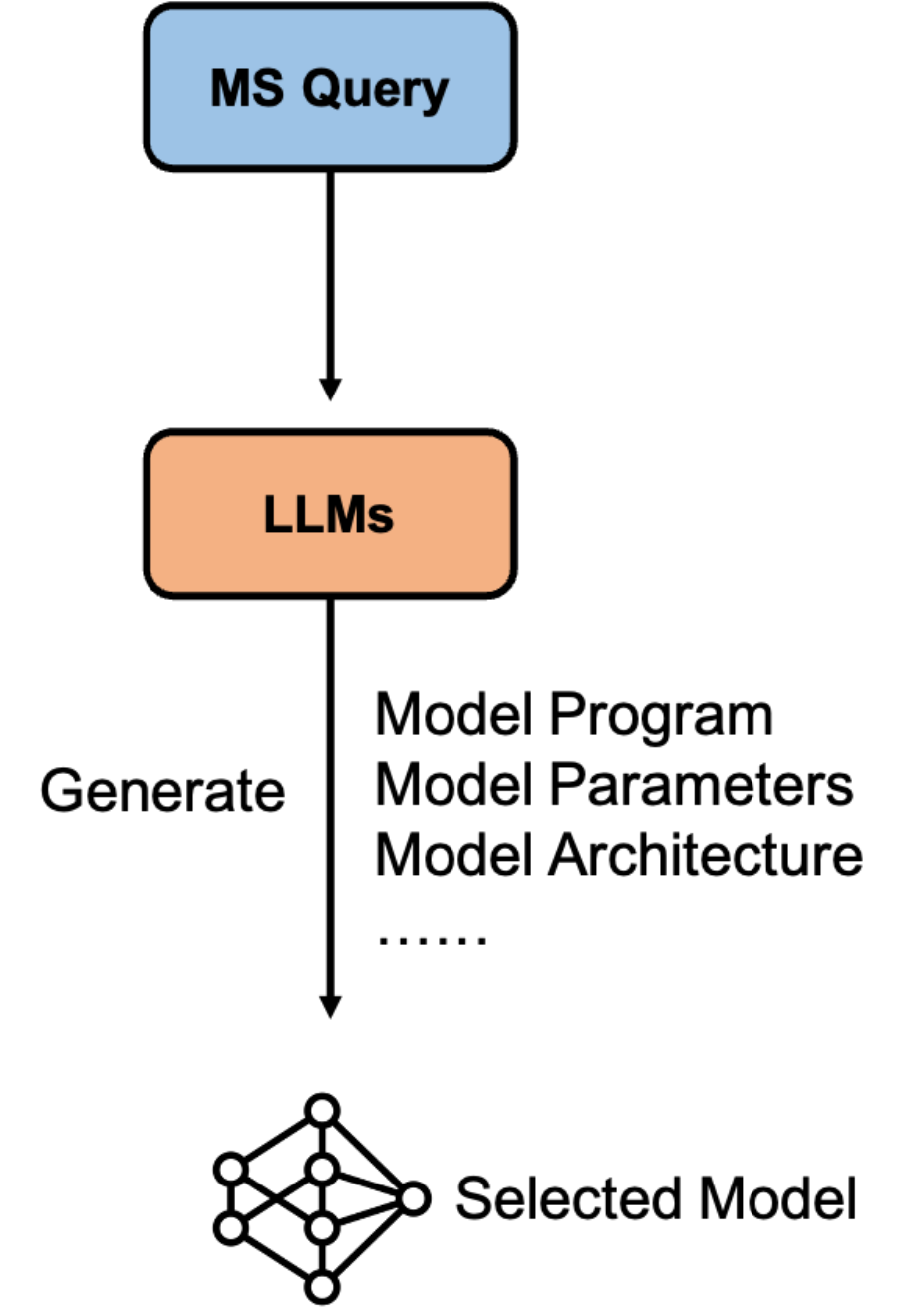}
    \label{fig-MS_ge}
    }
    \caption{The illustration of two distinct approaches for LLM-assisted Model Selection methods.}
    \label{fig-MS}
\end{figure}

\subsubsection{Retrieval-based Model Selection}

The core idea of retrieval-based methods is to first construct a repository of candidate models and then use LLMs to select the most suitable models based on task and model descriptions. In AutoMMLab \cite{yang2024autommlab}, the authors construct a comprehensive model zoo, where each model is paired with a detailed model card and pre-trained weights, taking AI safety concerns into account. Each model card includes attributes such as the model’s name, structure, parameters, floating point operations per second (FLOPs), inference speed, and performance metrics. An elaborate pipeline is then designed to automatically select the most appropriate model from the zoo, based on model performance and fuzzy matching scores between the model structure and the user’s specified requirements.

To address the context length limitations when processing model information, several approaches have been proposed. The multimodal framework AutoM$^3$L \cite{luo2024autom3l} catalogues candidate models in a model zoo, utilizing LLM-powered tools like ChatPaper \cite{ChatPaper} to automatically generate model cards, eliminating the need for manual documentation. Based on user directives, the system queries the top-5 models that match each input modality by comparing the text-based similarity between the embeddings of user requirements and the model card descriptions. From this refined subset, the Model Selection-LLM (MS-LLM) selects the most appropriate model. Similarly, AutoML-GPT \cite{zhang2023automl} and HuggingGPT \cite{shen2024hugginggpt} employ an in-context task-model assignment mechanism to dynamically select models. This mechanism first filters models that align with the task type, then selects the top-k models based on the number of downloads, from which LLMs choose the most suitable model based on the user’s query and task information provided in the prompt.

Notably, the model repositories in these approaches rely on explicit historical experience, which is critical for solving complex ML tasks but can be challenging due to the heterogeneous nature of the data (e.g., code, configurations, logs). To address this, MLCopilot \cite{zhang2023mlcopilot} introduces a two-stage knowledge-based reasoning approach that leverages LLMs to reason and solve tasks by drawing on knowledge from past experiences. In the offline stage, MLCopilot standardizes historical data into a unified format, creating experience and knowledge pools with the aid of LLMs. In the online stage, it retrieves relevant experiences and knowledge from these pools based on a novel task description, then interacts with LLMs to generate multiple ML solutions tailored to the task.

Additionally, the Model Agent within the AutoML-Agent framework \cite{trirat2024automl} performs model retrieval and hyperparameter suggestion, guided by insights provided by an Agent Manager regarding high-performing models and suitable hyperparameters for the specific ML task. Moreover, \cite{ozturk2022zero} introduces a zero-shot AutoDL method that meta-learns a large pre-trained model to select the best deep learning model from candidates, conditioned on dataset meta-features.

\subsubsection{Generation-based Model Selection}
Another class of methods leverages the powerful generative capabilities and vast ML/AI domain knowledge of LLMs. These approaches argue that relying on pre-defined models for a wide range of tasks is often impractical \cite{hollmann2024large, xu2024large}, particularly given the diverse requirements and characteristics of different ML tasks, such as dataset structures, feature types, and desired output formats. To tackle this challenge, \cite{xu2024large} proposes an end-to-end ML program synthesis approach that fully automates the generation and optimization of code across the entire ML workflow. Their Contextual Modular Generation framework breaks down the workflow into smaller modules, including model selection, with each module being generated independently by LLMs. Additionally, unit testing is employed to ensure compatibility between newly generated modules and pre-written, less variable optimization modules.

Other approaches focus specifically on the generation and optimization of model parameters. For instance, \cite{xiao2024verbalized} introduces the Verbalized Machine Learning (VML) framework, where the ML model is described and refined using natural language. In VML, the LLM's input text prompts act as the model parameters to be learned. Meanwhile, an optimizer—also parameterized by LLMs—dynamically updates the model parameters using current values, training data batches, and loss functions. This interaction between the learner and optimizer LLMs enables iterative model refinement, facilitating automatic model selection. In a similar vein, ModelGPT \cite{tang2024modelgpt} offers a framework that translates user data and descriptions into model parameters. This workflow consists of two key components: the Requirement Generator and the Model Customizer. The Requirement Generator utilizes LLMs to process user input, summarizing the task, analyzing data patterns, and condensing this information into a concise user requirement. This requirement is then processed by the Model Customizer, which determines the target model's architecture. Finally, the Model Customizer encodes the user requirement into a latent variable, which is decoded into model parameters using LoRA-assisted \cite{hu2021lora} Module-Wise hypernetworks, creating a customized model ready for predictive tasks. This represents a significant step towards AGI, enabling the creation of tailored AI models with minimal data, time, and expertise.

In graph learning, GL-Agent \cite{wei2024versatilegraphlearningapproach} illustrates another generation-based approach, focusing on neural architecture search (NAS). In this framework, LLM-based agents are responsible for configuring both the search space and the search algorithm. These agents leverage their extensive domain knowledge to select appropriate operation modules, matching them to specific learning tasks. By referencing resources like PyG documentation, the agents prepare candidate operations for each module, constructing a search space that aligns with hardware constraints—much like a human expert would. Once the search space is defined, the agent chooses a suitable neural architecture search algorithm based on summarized requirements, effectively guiding the model selection process.

\vspace{2em}
While LLMs offer considerable advantages in automating the model selection process, several challenges and limitations persist. In \textbf{retrieval-based} methods, the dependence on pre-built model repositories limits flexibility, especially when addressing novel or highly specialized tasks. These methods are constrained by the models available in the repository, which may not always be suited to the task at hand. Additionally, the effectiveness of these approaches relies heavily on comprehensive, well-maintained model cards and documentation, which can be labor-intensive to curate and difficult to keep up to date. Managing the context length limitations within LLM prompts also presents a challenge, particularly when dealing with large volumes of model information, which can diminish the quality of the recommendations.

In \textbf{generation-based} methods, while LLMs offer the flexibility to create customized models, a key challenge arises from the inherent stochasticity of LLM inference. The randomness in the generation process can lead to variability in the models selected, often resulting in inconsistent or suboptimal choices. Although ModelGPT \cite{tang2024modelgpt} constrains the candidate model architectures for specific tasks, determining their hyperparameters and parameters still requires a complex and sophisticated process. Additionally, LLMs often struggle with numerical precision \cite{yuan2023well}. Even when symbolic expressions are correctly interpreted, numerical errors during inference on specific input values can lead to fitting errors, compromising the accuracy of model selection. While LLMs are capable of generating advanced models, ensuring seamless compatibility between newly generated modules and pre-existing workflow components frequently requires manual intervention or extensive testing, diminishing the overall level of automation.

Across \textbf{both} approaches, addressing domain-specific constraints and maintaining the interpretability of selected or generated models remain significant hurdles. In addition, all LLM-driven methods face the broader risk of "hallucination," where LLMs generate plausible but incorrect models or configurations. This underscores the need for robust validation mechanisms and human oversight to ensure that the selected or generated models are both reliable and effective for their intended tasks.

\subsection{Hyperparameter Optimization}

Hyperparameter optimization (HPO), or hyperparameter tuning, is the process of selecting the optimal hyperparameters that enhance a machine learning model’s performance. Hyperparameters are configuration parameters set prior to training—such as the maximum depth in decision trees or learning rates in neural networks—and are not learned during training itself. The objective of HPO is to automatically identify the hyperparameter values that maximize a model’s performance on a given task \cite{bergstra2012random}. NAS is a specific instance of HPO, focusing on optimizing both the architecture and hyperparameters of neural networks simultaneously \cite{jawahar2023llm, zheng2023can}.

Traditional HPO methods like grid search, random search, and Bayesian optimization have been effective \cite{eggensperger2021hpobench}, but they are often computationally intensive and struggle to scale efficiently for large, complex models. Recently, LLMs have emerged as promising tools to enhance the efficiency and effectiveness of hyperparameter optimization by leveraging their contextual understanding and prior knowledge to propose optimized hyperparameter configurations. Based on whether actual training and testing are performed for hyperparameter settings, LLM-assisted HPO methods can be broadly categorized into two types: execution-based and prediction-based HPO.

\begin{figure}[htbp]
  \centering
  \includegraphics[width=0.7\textwidth]{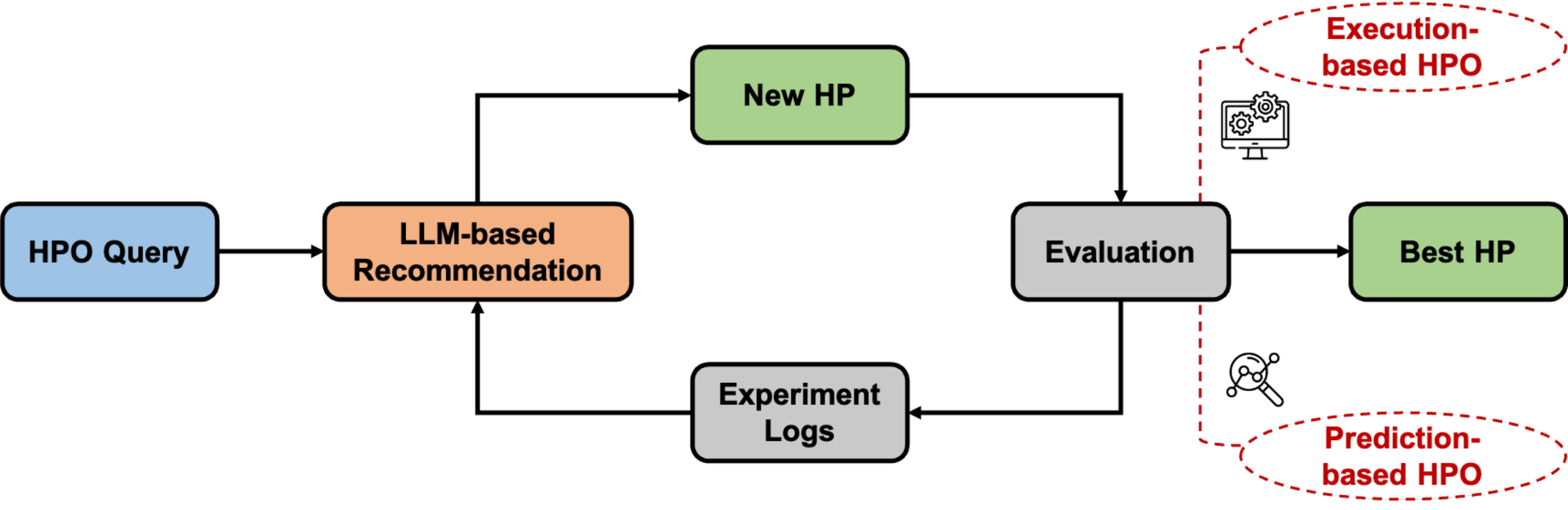}
  \caption{The overview of two categories of LLM-assisted Hyperparameter Optimization methods.}
  \label{fig-HPO}
\end{figure}

\subsubsection{Execution-based HPO}
Execution-based HPO refers to the process where hyperparameter settings recommended by LLMs are tested through actual training and evaluation of models on real machines. In the AutoMMLab system \cite{yang2024autommlab}, researchers introduce a “black-box” hyperparameter optimizer, called HPO-LLaMA, which is built by fine-tuning LLaMA-7B \cite{touvron2023llama}. HPO-LLaMA begins by receiving a general specification of the HPO problem, alongside request-specific details about the dataset and model. Based on this, it generates an initial hyperparameter configuration tailored to the task. After training the model with this configuration, HPO-LLaMA evaluates its performance on the test data and iteratively recommends refined hyperparameter settings based on the results, thereby optimizing performance. Similarly, GENIUS \cite{zheng2023can} also operates through an iterative refinement process to maximise a given performance objective of a NAS problem by using GPT-4 as a “black box” optimiser.

To address the challenge of the vast search space in neural architecture design, \cite{yu2023gpt} introduces GPT-NAS, a novel architecture search algorithm that combines a Generative Pre-Trained (GPT) model with an evolutionary algorithm (EA) as its search strategy. GPT-NAS operates on the assumption that a generative model pre-trained on a large-scale corpus can learn the underlying principles of constructing neural architectures. Leveraging this, GPT-NAS uses the GPT model to propose and reconstruct reasonable architecture components given the basic one and then utilizes EAs to search for the optimal solution. This approach significantly reduces the search space and enhances efficiency by incorporating prior knowledge and rapidly eliminating low-quality architectures during the search process. Likewise, to improve the efficiency and reduce manual intervention in EA-based NAS, \cite{morris2024llm} introduces “Guided Evolution” (GE), a novel framework that combines LLMs' human-like reasoning with the robustness of NAS through genetic algorithms. The key innovation, termed “Evolution of Thought” (EoT), allows LLMs to iteratively refine both model architectures and hyperparameters based on feedback from previous iterations. By leveraging training results, LLMs can make informed adjustments through techniques such as code segment mating and mutation, ultimately enhancing both model accuracy and efficiency.

Moreover, AutoM$^3$L \cite{luo2024autom3l} introduces HyperParameter Optimization-LLM (HPO-LLM), which collaborates with external tools like ray.tune to optimize hyperparameters. LLMs, leveraging their vast knowledge of machine learning training processes, first generate comprehensive descriptions for each hyperparameter in the configuration file. These descriptions, along with the original configuration, form the prompt for HPO-LLM, which then recommends optimal hyperparameter settings. The recommended hyperparameters and their corresponding search intervals are incorporated into the ray.tune search space for optimization, reducing the need for manual specification by users.

\subsubsection{Prediction-based HPO}
Prediction-based HPO refers to methods where LLMs recommend hyperparameter settings based on predicted performance, bypassing the need for actual model training and evaluation \cite{trirat2024automl}. This approach allows for faster and more resource-efficient optimization. An example of such a system is AgentHPO \cite{liu2024large}, a two-agent framework consisting of a Creator and Executor agent, which leverages LLM-powered autonomous agents to overcome the complexities of traditional AutoML methods. The Creator agent initiates the optimization process by allowing users to input task-specific details in natural language. It then interprets the input and generates initial hyperparameters, mimicking the expertise of a human specialist. The Executor agent takes the hyperparameters generated by the Creator and handles tasks like training models, recording experimental results, and conducting outcome analysis. The Creator agent further iteratively refines the hyperparameters based on the insights gleaned from the Executor’s training history, streamlining the optimization process and making it more intuitive and efficient. Furthermore, AutoML-GPT \cite{zhang2023automl} predicts model performance by generating a training log for a given hyperparameter configuration based on the provided data card and model card. The system then tunes the hyperparameters by analyzing the predicted training log, which includes relevant metrics and information from the simulated training process.

Bayesian optimization (BO) is a widely-used approach for optimizing complex and costly black-box functions, especially in hyperparameter tuning \cite{jones1998efficient}. However, its success hinges on efficiently balancing exploration and exploitation. To address this challenge, LLAMBO \cite{liu2024largelanguagemodelsenhance} frames the BO problem in natural language, allowing LLMs to iteratively propose and evaluate promising solutions based on historical evaluations. LLMs can initiate the BO process via zero-shot warm-starting, sampling candidate points from regions with high potential based on past observations and the problem description, and evaluating these candidate points through a surrogate model, thus enhancing the overall efficiency of the optimization process.

\vspace{2em}
In summary, Execution-based and Prediction-based HPO methods each offer distinct advantages and limitations depending on the machine learning task at hand. \textbf{Execution-based} HPO relies on actual training and evaluation, making it highly reliable in assessing the impact of hyperparameter settings on model performance. In this context, LLMs function as intelligent optimizers, utilizing their vast knowledge to propose hyperparameters that align with task requirements and real-time feedback from model training. However, this method is resource-intensive, demanding substantial computational power and time, especially when dealing with large-scale models or datasets. While execution-based approaches ensure precision, their scalability is often constrained by the complexity of the search space and the model architecture.

Conversely, \textbf{Prediction-based} HPO leverages LLMs not only to recommend hyperparameters but also to predict model performance without full training cycles for each configuration. Systems like AgentHPO \cite{liu2024large} accelerate the process by simulating training outcomes and hyperparameter effects using LLM-generated logs or prior task knowledge, making these methods more resource-efficient and faster. However, prediction-based approaches face significant challenges related to accuracy. Relying on LLM-generated predictions can introduce biases or errors, especially for complex tasks or unfamiliar domains. While methods like LLAMBO \cite{liu2024largelanguagemodelsenhance} attempt to mitigate these issues by leveraging prior evaluations, predictive uncertainties can still lead to suboptimal configurations, particularly when nuanced interactions between hyperparameters are involved.

\section{LLMs for Workflow Evaluation}
\label{5}

After the ML model is generated, evaluating workflow performance becomes an essential step.  The conventional approach is to train the model to convergence and assess its performance on a validation dataset. However, this method can be time-consuming and computationally expensive, especially for large-scale datasets and complex workflows \cite{lazebnik2022substrat, zhang2023using}. For instance, NASNet required 500 GPUs over four days, consuming 2,000 GPU hours to complete a search for the CIFAR-10 classification task \cite{zoph2018learning}. Given these resource demands, LLM-driven methods have emerged as a promising way to expedite workflow evaluation. Notably, evaluation is intrinsically linked to HPO, as hyperparameter tuning frequently relies on performance metrics derived from evaluation. Consequently, there is some overlap between methods applied to both processes.

In systems like AutoML-GPT \cite{zhang2023automl}, LLMs are used to predict key metrics like loss and accuracy by generating simulated training logs based on specific hyperparameter settings, data cards, and model cards. However, these predictions are based on the overall dataset, providing a coarse-grained estimation of model performance. More fine-grained evaluation is found in the VML system \cite{xiao2024verbalized}, where a learner agent, parameterized by an LLM prompt, predicts the output of individual inputs one-by-one. These predictions are compared with target outputs by an optimizer agent, enabling a more granular comparison of the model's performance across various instances.

Additionally, some approaches utilize large pre-trained models to predict workflow performance. In the optimization phase of Text-to-ML \cite{xu2024large}, zero-cost (ZC) proxies \cite{mellor2021neural} are incorporated to estimate the performance of candidate ML programs using at most a single forward/backward propagation pass, drastically reducing evaluation costs. Another example is TabPFN \cite{hollmann2022tabpfn}, a pre-trained Transformer model designed to solve small tabular classification problems in just seconds. TabPFN is trained offline to approximate Bayesian inference on synthetic datasets generated from a prior, enabling it to efficiently process training data and generate predictions in a single forward pass.

LLMs offer several advantages in workflow evaluation. First, they significantly reduce time and computational resources by simulating training processes and predicting performance metrics without the need to run models on physical machines. This is particularly beneficial for large-scale workflows or models that would otherwise require extensive hardware resources. Additionally, LLMs can provide early-stage performance predictions based on hyperparameter settings, improving the efficiency of model selection and optimization processes. Their ability to offer fine-grained evaluation, such as predicting individual output results, further enhances the accuracy of model assessment. 

However, LLM-driven workflow evaluation also comes with challenges. The generalizability of LLM predictions may be limited if the pre-training data does not align well with the specific task, leading to inaccurate performance estimates and potentially suboptimal optimization decisions. LLMs may also face difficulties with complex or large-scale workflows, where nuanced interactions between model components are harder to predict. Moreover, LLMs are susceptible to hallucinations, which can result in plausible-sounding but incorrect performance metrics, potentially undermining the reliability of evaluations.

\section{Open Challenges and Future Directions}
\label{6}

In addition to the many promising ways LLMs can be leveraged for constructing and optimizing ML workflows, their integration also presents certain risks. While the earlier sections focused on challenges and limitations unique to each workflow component, this section provides a broader summary of the common challenges faced throughout the entire process and proposes potential solutions to mitigate these issues. Furthermore, we outline key directions for future research aimed at enhancing the effective integration of LLMs into ML workflow construction and optimization.

\subsection{Open Challenges}

\subsubsection{Potential Data Leakage}
One of the major concerns when integrating LLMs into ML workflows is the risk of data leakage, commonly referred to as data snooping. Since LLMs are pre-trained on vast amounts of publicly available data, including many common ML datasets, this overlap can lead to biased evaluations. LLMs may "memorize" portions of training or test datasets, resulting in inflated performance estimates during evaluation. Additionally, the pre-training data from the internet may embed biases, leading to biased outputs during various components of the ML workflow, such as feature selection \cite{jeong2024llm} and feature synthesis \cite{hollmann2024large}. These biases can manifest as gender or racial disparities, potentially impacting model fairness and decision-making across ML tasks \cite{luo2024autom3l}.

To mitigate these issues, three primary solutions have been proposed. First, researchers can use datasets that were generated after the LLM’s pre-training phase or private datasets that the LLM has not been exposed to \cite{zhang2023mlcopilot}. However, this solution raises challenges, such as ensuring access to realistic and publicly available datasets for fair benchmarking across different AutoML systems. Second, fine-tuning the LLM to remove knowledge of specific datasets could reduce the impact of data leakage. However, existing fine-tuning techniques are limited in their ability to fully erase prior knowledge, and identifying diverse and inclusive fine-tuning data that encompasses various dimensions like gender, race, and culture remains challenging \cite{tornede2023automl, luo2024autom3l}. Third, implementing robust user-involved evaluation protocols to account for potential data leakage has been proposed as a practical solution. For example, incorporating interactive post-processing steps or rule-based frameworks can help ensure that the generated results are free from harmful biases \cite{luo2024autom3l}. Additionally, refining prompts by including diverse examples or emphasizing fairness criteria can guide the LLM in making unbiased decisions and improve the reliability of workflow outputs \cite{hollmann2024large}.

\subsubsection{Complicated Prompt Engineering}
The effectiveness of LLMs heavily depends on the quality and structure of prompts, which must clearly communicate the task, data, and model requirements—a process referred to as prompt engineering \cite{chen2023unleashing}. However, several challenges complicate this process.

First, context length limitations pose a significant challenge in LLM-based workflows. LLMs are constrained by the maximum number of tokens they can process in a single prompt, which restricts their ability to handle large datasets, complex task descriptions, and intricate ML workflows \cite{shen2024hugginggpt}. For instance, in VML \cite{xiao2024verbalized}, the LLM's context window limits the dimensionality of input data and the batch size, preventing the system from processing high-dimensional data or optimizing models efficiently. While advancements in LLMs have extended token limits \cite{islam2024gpt}, future work on long-context LLMs equipped with memory mechanisms and positional interpolation may further increase token capacity, enabling the seamless integration of large datasets, multiple models, and task details into a single workflow \cite{ding2024longrope, wang2024beyond}.

Second, heterogeneous data and task requirements complicate prompt design. Prompts must encode not only raw data but also model configurations, optimization strategies, and evaluation metrics, which adds layers of complexity \cite{zhang2023mlcopilot, hollmann2024large}. Ensuring that these prompts produce robust and accurate outputs can be difficult, particularly in workflows with varying data types and requirements. Recent research has proposed multi-step refinement frameworks, where LLMs iteratively improve their outputs through user feedback, using techniques like reflection and multi-round dialogues to enhance the reliability of the generated results \cite{wei2024versatilegraphlearningapproach}. Additionally, users can enhance prompt design by leveraging detailed logging mechanisms that trace LLM decision-making processes, providing transparency and insights into how LLMs select components for the workflow.

\subsubsection{Hallucination}
Hallucination in LLMs refers to instances where the generated content deviates from either real-world facts or the user’s instructions. There are two primary forms of hallucinations: factuality hallucinations and faithfulness hallucinations \cite{huang2023survey}, both of which pose significant risks to the integrity and reliability of ML workflow construction.

Factuality hallucinations occur when the generated content contradicts real-world facts, resulting in factual inconsistencies or fabrications. In systems like MLCopilot \cite{zhang2023mlcopilot}, these hallucinations can appear when LLMs recommend incorrect or outdated models and hyperparameters, which may seem plausible but are inappropriate for the task at hand. Similarly, VML \cite{xiao2024verbalized} demonstrates how even correct symbolic interpretations by the LLM can lead to numerical errors \cite{yuan2023well}, causing inaccurate predictions during model inference. In CAAFE \cite{hollmann2024large}, factuality hallucinations are evident in the feature generation process, where LLMs generate features that appear logically sound but are not grounded in the actual data distribution, resulting in distorted model performance and user misinterpretation.

On the other hand, faithfulness hallucinations occur when the generated content diverges from the user’s instructions or input context, or fails to maintain self-consistency. These hallucinations can be further divided into instruction inconsistency, context inconsistency, and logical inconsistency \cite{huang2023survey}. For example, in MLAgentBench \cite{huang2024mlagentbench}, LLM-powered agents sometimes modify ML pipelines in ways that contradict the user’s original task specifications, leading to instruction inconsistencies. HuggingGPT \cite{shen2024hugginggpt} probably exhibits both context and logical inconsistencies, where the outputs generated by the LLM fail to align with the task context or maintain internal logical coherence, leading to unreliable and unpredictable results.

To mitigate these hallucinations, future research should focus on integrating knowledge graphs or validated data sources to strengthen the factual grounding of LLM-generated content \cite{pan2024unifying}, thereby reducing the likelihood of factuality hallucinations. Additionally, employing ensemble models that cross-validate LLM outputs before final recommendations can help mitigate the risks of factual fabrications. To address faithfulness hallucinations, incorporating contextual validation mechanisms can ensure that LLM-generated outputs remain consistent with user instructions and input contexts. Finally, RLHF technique could be employed to iteratively fine-tune LLM responses, improving alignment with user expectations and reducing inconsistencies. By implementing these solutions, the reliability and trustworthiness of LLM-driven ML workflows could be significantly enhanced, especially in critical applications where hallucinations could have serious consequences. Overall, it is crucial to anticipate the possibility of erroneous outputs when integrating LLMs into ML workflow construction and optimization, and to develop probabilistic methods to effectively detect and mitigate these risks.

\subsubsection{Interpretability}
Interpretability is a vital aspect of ML workflows, especially those involving complex models or automated systems like AutoML \cite{drozdal2020trust}. Ensuring that both experts and non-experts can understand how decisions are made by systems is crucial for building trust, improving decision-making, and maintaining accountability. Many current systems address this need by providing transparency through logging and tracing, as seen in Text-to-ML \cite{xu2024large}, where prompts and responses are recorded to help users trace decisions. Similarly, AgentHPO \cite{liu2024large} enhances interpretability by logging experimental results and hyperparameter adjustments, allowing users to track model evolution and performance.

Despite these improvements, significant challenges remain in ensuring interpretability for LLM-driven workflows. Over-reliance on textual explanations, for example, can create a false sense of security among users, particularly those less familiar with ML intricacies. Users might assume the model is fully optimized when, in reality, it may make implicit assumptions or overlook ambiguities in the input data or task description. Additionally, the opaque nature of LLMs, which rely on vast pre-trained knowledge, can make it difficult to fully understand the rationale behind certain model or hyperparameter recommendations, thereby reducing transparency \cite{xu2024large}.

To address these challenges, future research should focus on integrating explainable AI (XAI) techniques into LLM-driven workflows. XAI methods \cite{hoffman2018metrics} could offer deeper insights by complementing textual explanations with visual tools, causal diagrams, or interactive features that highlight how specific features or parameters affect model performance. Furthermore, incorporating human-in-the-loop mechanisms \cite{hollmann2024large, pmlr-v235-lindauer24a} would ensure that users are not merely passive recipients of information but actively engaged in confirming the validity and appropriateness of the system's decisions.

\subsubsection{Resource Consumption}
While LLM-driven ML workflows offer substantial potential for automation and optimization, they also introduce significant challenges in terms of resource consumption. The computational demands of LLMs, particularly during inference, can be considerably higher than those of traditional machine learning methods. Although some studies suggest that the resource impact during certain workflow phases is minimal \cite{luo2024autom3l}, the overall costs in terms of human labor, time, money, and computational power remain considerable.

For example, LLAMBO \cite{liu2024largelanguagemodelsenhance} shows that even without fine-tuning, LLM inference incurs a significantly larger computational footprint compared to traditional Bayesian Optimization (BO) algorithms. Similarly, AutoM3L \cite{luo2024autom3l} highlights that while LLMs are primarily utilized during the setup phase of an ML pipeline, the bulk of computational resources are consumed during the actual training phase. Though the cost introduced by LLMs during setup may seem minimal, these costs accumulate across the entire pipeline, particularly due to real-time API calls and the need for scalable infrastructure to support continuous operations.

Addressing these challenges requires future research to focus on improving the computational efficiency of LLM-driven workflows. One potential direction is the development of lightweight LLM architectures specifically tailored for ML tasks, reducing the computational load without sacrificing performance. Additionally, hybrid models that combine the strengths of traditional algorithms with LLMs, as demonstrated by LLAMBO's fusion with more efficient methods, could further enhance resource efficiency. Finally, optimizing the orchestration of LLM interactions, as suggested by HuggingGPT \cite{shen2024hugginggpt}, can reduce redundant steps, cut down on processing time, and improve real-time performance. Embracing these strategies, in line with the principles of Green AutoML \cite{tornede2023towards}, will be essential to ensure that the advantages of LLM-driven workflows are realized without imposing excessive resource demands.

\subsubsection{Social Impact}
The integration of LLM-driven ML workflows offers numerous societal benefits but also raises significant concerns, particularly regarding the displacement of human labor and ethical considerations. On the positive side, LLM-based workflows have the potential to democratize ML by reducing the level of expertise required to develop and deploy models. Tools like Text-to-ML \cite{xu2024large} and AutoMMLab \cite{yang2024autommlab} automate many ML tasks, making the technology accessible to a broader audience. This democratization can empower individuals and organizations previously excluded from the field due to resource constraints or lack of specialized knowledge, fostering innovation across various domains.

However, this automation also presents challenges. One of the key concerns is the potential displacement of human workers. As systems become more autonomous and capable of managing complex ML tasks, there is a risk that they could diminish or replace roles traditionally held by data scientists, engineers, and researchers. This shift could have significant implications for employment within the field of ML and beyond \cite{huang2024mlagentbench}. For instance, CAAFE \cite{hollmann2024large} illustrates how automating routine data science tasks could lead to a reduction in the demand for manual labor, potentially displacing workers in data-centric industries. Nevertheless, it also emphasizes that such automation could free up human professionals to focus on more strategic decision-making, enhancing their productivity. Moving forward, developing frameworks where AI augments rather than replaces human labor could mitigate these concerns by encouraging a human-in-the-loop approach to maintain human oversight and creativity \cite{huang2024mlagentbench}.

Ethical considerations are another critical issue. LLMs, especially in highly automated workflows, could generate unethical, biased, or even harmful outputs when exposed to biased training data or malicious user input. For example, MLCopilot \cite{zhang2023mlcopilot} warns that while LLMs operate within user-defined boundaries, when applied in more complex or less-constrained scenarios, there is a risk of generating unpredictable or unethical outcomes. AutoMMLab also acknowledges the potential for LLM-generated models to produce toxic or offensive content, particularly when exposed to harmful inputs \cite{yang2024autommlab}. To mitigate these risks, future research should focus on developing mechanisms for continuous monitoring of LLM outputs, ethical auditing frameworks, and incorporating fairness and bias mitigation techniques at every stage of the workflow. Moreover, impact assessments should be conducted before deploying these systems in sensitive applications, ensuring that both ethical standards and societal impacts are properly considered \cite{xu2024large}.

\subsection{Future Directions}
Building on the challenges and opportunities discussed, future research directions present exciting possibilities for advancing LLM-driven ML workflows. 

One promising direction is the development of \textbf{end-to-end ML workflow construction using LLMs}, where LLMs autonomously handle every stage of the ML pipeline—from data preprocessing to model evaluation—creating fully automated workflows. The goal is to further reduce human intervention, enabling seamless construction, optimization, and deployment of ML models. Some existing agent-based methods, such as AIDE \cite{wecoai_wecoaiaideml_2024}, ResearchAgent \cite{huang2024mlagentbench}, CodeActAgent \cite{wang2024executable, wang2024openhands}, and DS-Agent \cite{guo2024ds}, already offer implicit stage-wise solutions for end-to-end ML workflow generation. These approaches leverage the generative capabilities of LLMs to produce solution drafts and iteratively refine them based on performance feedback and interactions with the environment. However, this process often lacks transparency, making it difficult for users to personalize or modify individual workflow components. In contrast, explicit end-to-end ML workflow generation could provide a complete ML solution while allowing users to control and customize each component.

Another promising direction involves \textbf{integrating LLMs with specialized models} to enhance workflow generation. Instead of relying solely on LLMs to handle the entire process, hybrid systems can capitalize on the strengths of LLMs for tasks such as data preprocessing and feature selection, while leveraging domain-specific models for more specialized functions like hyperparameter optimization or model evaluation. This hybrid approach combines the flexibility and broad capabilities of general large models with the efficiency and domain expertise of tailored small models, leading to more robust and scalable ML workflows that can adapt to diverse tasks and requirements.

\section{Conclusion}
\label{7}
This paper provides a detailed and systematic review of how LLMs contribute to constructing and optimizing ML workflows across different stages of the ML pipeline (Fig. \ref{fig-MLWorkflow}). From data and feature engineering to model selection and hyperparameter optimization, and workflow evaluation, LLMs bring diverse capabilities that can streamline and enhance various steps in the workflow. Despite the significant advancements LLMs have made in automating and improving ML workflows, many challenges remain, and we are still in the early stages of addressing these complexities. This review not only highlights the current achievements but also outlines several open problems and suggests important future research directions.

The potential of LLMs to revolutionize ML workflows is immense, and with continued research and collaboration within the community, the existing challenges can be overcome. We are confident that the innovations and opportunities provided by LLMs will lead to breakthroughs that transform the development and application of ML/AI. By offering a comprehensive understanding of the role of LLMs in ML workflows, this review will be a valuable resource for researchers and practitioners entering this evolving field, guiding future advancements and fostering continued progress.

\bibliographystyle{unsrtnat}
\bibliography{references}  






\end{document}